\documentclass[conference]{IEEEtran}
\IEEEoverridecommandlockouts
\usepackage{cite}
\usepackage{xcolor}
\usepackage{amsmath,amssymb,amsfonts}
\usepackage{algorithmic}
\usepackage{graphicx}
\usepackage{textcomp}
\usepackage{xcolor}
\def\BibTeX{{\rm B\kern-.05em{\sc i\kern-.025em b}\kern-.08em
    T\kern-.1667em\lower.7ex\hbox{E}\kern-.125emX}}
\usepackage{enumitem}
\usepackage[utf8]{inputenc}
\begin{document}
\pagenumbering{arabic}
\setcounter{page}{1}
\title{On Diversity in Discriminative Neural Networks\\}

\author{\IEEEauthorblockN{ Brahim Oubaha}
\IEEEauthorblockA{\textit{	Mathematical and Electrical Engineering} \\
\textit{IMT Atlantique}\\
Brest, France \\
brahim.oubaha@imt-atlantique.fr}

\and
\IEEEauthorblockN{ Claude Berrou}
\IEEEauthorblockA{\textit{	Mathematical and Electrical Engineering} \\
\textit{IMT Atlantique}\\
Brest, France \\
claude.berrou@imt-atlantique.fr}
\and
\IEEEauthorblockN{ Xueyao Ji}
\IEEEauthorblockA{\textit{Center of Brain Sciences} \\
\textit{Institute of Basic Medical Sciences}\\
Beijing, China \\
xy.ji@foxmail.com}
\and

\IEEEauthorblockN{ Yehya Nasser}
\IEEEauthorblockA{\textit{	Mathematical and Electrical Engineering} \\
\textit{IMT Atlantique}\\
Brest, France \\
yehya.nasser@imt-atlantique.fr}
\and

\IEEEauthorblockN{ Raphaël Le Bidan}
\IEEEauthorblockA{\textit{	Mathematical and Electrical Engineering} \\
\textit{IMT Atlantique}\\
Brest, France \\
raphael.lebidan@imt-atlantique.fr}
\and

}

\maketitle

\begin{abstract}

Diversity is a concept of prime importance in almost all disciplines based on information processing. In telecommunications, for example, spatial, temporal, and frequency diversity, as well as redundant coding, are fundamental concepts that have enabled the design of extremely efficient systems. In machine learning, in particular with neural networks, diversity is not always a concept that is emphasized or at least clearly identified. This paper proposes a neural network architecture that builds upon various diversity principles, some of them already known, others more original. Our architecture obtains remarkable results, with a record self-supervised learning accuracy of 99. 57\% in MNIST, and a top tier promising semi-supervised learning accuracy of 94.21\% in CIFAR-10 using only 25 labels per class.

\end{abstract}

\begin{IEEEkeywords}
Neural network, diversity, competition, sparsity, self- and semi-supervised learning, ensemble learning.
\end{IEEEkeywords}

\section{Introduction}

In the information sciences, the principle of diversity consists in combining information from different sources to better estimate the data. Diversity is all the more effective when the sources are decorrelated, i.e. when the information they provide is not processed in the same way and/or does not derive from the same observations. This ideal condition is rarely met, and we generally make do with partially correlated information.
It is probably the field of telecommunications that has benefited most from the principle of diversity in moving towards very high-performance systems, both fixed and mobile. In the time domain, channel coding (or error correcting coding) makes it possible to transmit the binary elements of an augmented (redundant) version of the original message at different times (and therefore generally subject to different disturbances) and to benefit from this redundancy in the receiver. In the frequency domain, techniques such as Orthogonal Frequency-Division Multiplexing can more or less eliminate spectrum irregularities and interferences. Pruning techniques may also be considered to remove inappropriate parts of the bandwidth. The spatial dimension is of course also used, with multi-path techniques such as Multiple-Input Multiple-Output taking advantage of the particular properties of the wave paths. Other types of diversity can be exploited at higher system levels (multiuser, multistandard, etc.).
In contrast, the classic architecture of a neural network, i.e. a few convolution layers followed by a classifier with a simple one-hot output (as many neurons as classes), does not reveal any deliberately introduced diversity technique. It could of course be pointed out that the totality of the weights of a neural network's connections is always oversized and therefore redundant. However, in the absence of a theory on neural network capacity and redundancy, we cannot really speak of intentional, controlled diversity.
Analogies can however be drawn between different types of diversity found in digital communications and in neural networks:

\subsection{Channel Coding}

Two techniques can be related to channel coding (redundant coding). The first, of high importance in self-supervised and semi-supervised applications, is data augmentation. This involves submitting several distorted versions (rotation, cropping, mirroring, etc.) of the same sample to the network. Redundancy rates are therefore several hundred percent. The second technique involves increasing the length of the network output by multiplying the number of neurons that must be activated for a given class. This is known as distributed coding. The redundancy rate is determined by the length of the output and can be several thousand percent. A theory of this process has been developed under the name of Error Correcting Output Coding (ECOC) \cite{b1}.

\subsection{Spatial Coding}
A convolution layer can be presented as a spatio-temporal encoding layer. This is because the implementation of filters seeking to extract features independently of coordinates involves sharing the synaptic weights of these filters. There is therefore both redundant coding (repeated weights) and spatial coding (the search for a certain invariance with respect to coordinates).
Regularization techniques such as dropout or drop-connect can also be assimilated to a form of spatial diversity.
Another type of spatial diversity, not often implemented to our knowledge, can be provided by the sparsity of connection matrices. This concept is developed in section II.

\subsection{Pruning}

A famous example of pruning in digital communications is Discrete MultiTone (DMT) modulation, which enabled the massive development of the Asymmetric Digital Subscriber Line (ADSL) application. This modulation divides the spectrum into multiple sub-channels whose capacity (number of bits transmitted per unit of time) is evaluated once and for all on a fixed channel (telephone pair). The least favorable sub-channels are assigned the lowest data rates. Some sub-channels may even be discarded. In a neural network, which will also eventually become a fixed device, pruning consists in removing the least discriminating paths with regard to the categories to be recognized. The analogy is relative, however, because in the first case, the aim is to maximize transmission throughput, whereas pruning in a neural network aims to simplify implementation and reduce computational requirements.

\subsection{Ensemble processing}
In the world of telecommunications, the most representative example of an ensemble processing is probably a constellation of satellites such as OneWeb or Starlink. In this type of system, the operation is unimodal, meaning that each satellite is entrusted with the task of communicating with the earth using the same transmission mode and the same type of equipment. The only parameter that distinguishes one satellite from another, towards a potential user, is the link budget, on which the choice of the most favorable satellite is based. In the field of discriminative neural networks, a unimodal ensemble processing does not consist in selecting one network among several, but in using all or almost all of them at the same time for the inference task \cite{b2}. Networks differ in the initialization of their weights or in their hyperparameters, which diversifies the ways in which they learn, particularly with regard to the inevitable local minima. Often, a simple majority vote decision or, if weighted decisions are available, a probabilistic vote is enough to improve performance compared to that of a single network. Ensemble processing can be performed independently by each network (in this case, it is better to call this ensemble inference rather than ensemble learning) or by linking their operations, using some information transfer algorithm \cite{b3}.

\section{Competition and Sparsity}

Competition has played an important role in the evolution of species, most often due to limited resources. The same is true for nearby neurons in the nervous system, because the energy provided by the local blood flow does not allow them all to activate at the same time. It is then not unreasonable to think that the brain relies on some kind of competition in order to learn and memorize while saving energy, and the same applies for artificial networks, by bio-inspired analogy.
The first paper to highlight the benefits of using the competition principle in neural networks dates back ten years \cite{b4}. It showed that the classical activation function of neurons in a hidden layer, usually a sigmoid or ReLU function, can be advantageously replaced by a Local Winner-Takes-All (LWTA) function in blocks of several neurons. In a given layer, this technique thus puts into competition all the activities coming from the neurons of the underlying layer through a full matrix of connections.
It turns out that this principle works just as well, and perhaps even better, if the matrix is sparse, or even very sparse, although this possibility was not considered in \cite{b4}. In this situation, the competition is performed between subpopulations of the underlying neurons, rather than within the whole population. These small groups of neurons are selected completely at random, since the drawing of the matrix of sparse connections is itself random. In addition to the diversity brought about by multiple, quasi-independent competitions within the considered layer, extra-diversity can be added when learning by an ensemble of networks. As each network is initialized in a different way (each time by a different seed that decides the topology of connections), the number of different competitions in the ensemble is increased.
The realities of competition (see \cite{b4} for relevant references) and sparse connections \cite{b5,b6} are considered proven and fundamental properties in mammalian cortexes. \emph{Our contribution appears to be the first to combine these two bio-inspired principles in artificial neural networks, and to evaluate the potential of such architectures for self- and semi-supervised learning}. We note that the competition and sparsity are very simple to implement. No particular pre- or post-processing is needed to make the best use of. We will only rely on the now classical functions adopted in neural networks, such as data augmentation, convolutional layers, batch normalization, pseudo-label estimation, etc.

\section{Proposed learning method}
\begin{figure*}[h] 
    \centering
    \includegraphics[width=\linewidth]{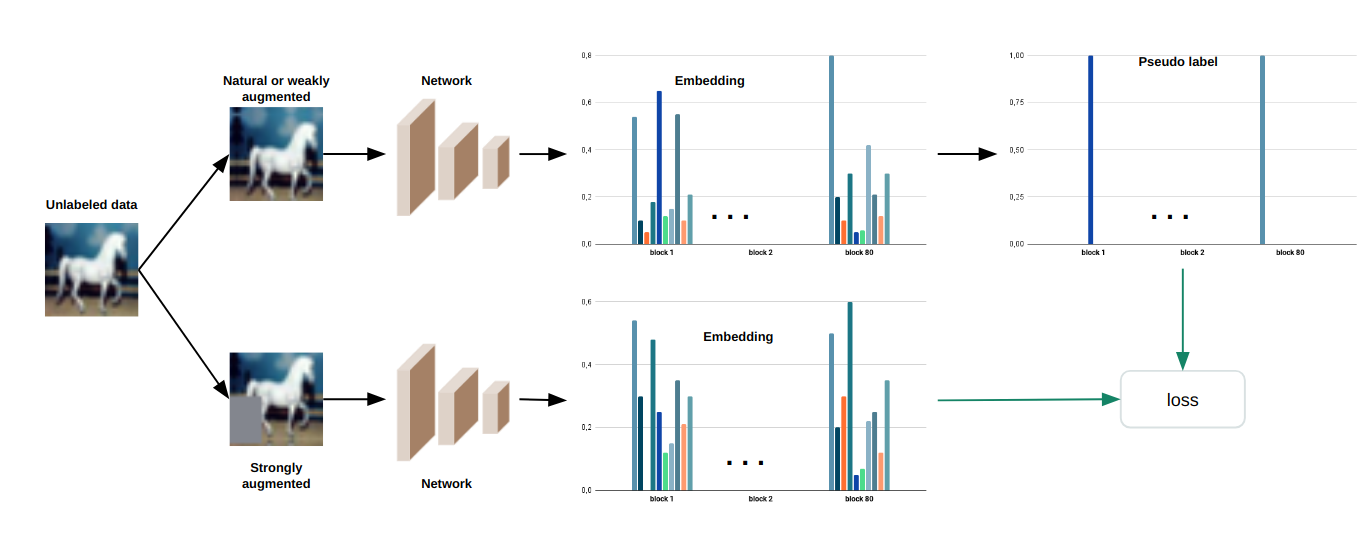}
    \caption{Overview of the proposed learning method for unlabeled data. For MNIST, the size of the blocks is 8 and 10 for CIFAR-10.}
    \label{fig:MethodOverview}
\end{figure*}

In this section, we start by revisiting preliminary work in self and semi-supervised learning frameworks, in particular those involving the principle of data augmentation and label estimation. Then we introduce the proposed learning method, detailing the architectural design of the models for each dataset. Next, we elaborate on our label estimation algorithm, highlighting its main features and operation.

Semi-supervised learning lies between supervised and unsupervised learning. It involves the use of a small amount of labeled data in conjunction with a large amount of unlabeled data during the training process. This approach is particularly beneficial in scenarios where labeled data is scarce or expensive to obtain, but there is an abundance of unlabeled data.

Consistency regularization, a key feature of many advanced semi-supervised learning algorithms, exploits unlabeled data based on the principle that a model should produce consistent predictions for different perturbed versions of the same image. This concept was initially introduced in an earlier work \cite{b7}, and has been more widely recognized in subsequent studies \cite{b8,b9}. It is implemented by training the model with traditional supervised classification loss and an additional loss function that handles unlabeled data, thus improving the model's ability to learn from a wider spectrum of data.

Another common approach in semi-supervised learning is pseudo-labeling \cite{b10}, where the model uses its predictions on unlabeled data to generate artificial labels. These pseudo-labels are then used in subsequent training to refine the model's performance.

Our semi-supervised learning approach leverages the strength of both labeled and unlabeled data, strategically incorporating data augmentation to enhance model performance. This allows us to make efficient use of all available data, combining the reliability of labeled examples with the broader coverage provided by unlabeled examples, thus improving the efficiency and robustness of the model.

Conversely, in our self-supervised learning approach, we exclusively rely on unlabeled data for training, omitting the use of labeled data. This method focuses on extracting meaningful patterns and structures from the data itself without direct guidance from explicit labels. 

To efficiently handle unlabeled data, our approach starts with pseudo-labeling, a critical step where we utilize our model on subtly altered data to generate pseudo-labels. This is instrumental in guiding the learning process, even without standard labeled data. What sets our label estimation process apart is its unique treatment of embeddings. Within each block of size \(n\) of the embedding space (embedding refers to the output of the model), we identify the maximum value and assign it a label of 1, while all other elements within the block are set to 0, following the principle of competition presented in Section II. This selective activation within the embedding space effectively highlights the most prominent features or characteristics captured by each block, acting as a form of dimensionality reduction and targeted feature amplification. 

Subsequently, we expose the same batch of data to strong deformations, such as extensive augmentation or distortion, while utilizing the estimated label as the target. This method effectively challenges the model to maintain its predictions under more significant variations, thereby enhancing its robustness and ability to generalize from complex or noisy data. The complete learning process is illustrated in Figure \ref{fig:MethodOverview}.

It is worth noting that for the labeled batch, nothing changes compared to other supervised learning approaches. Labeled data continue to serve as a reliable source of ground truth, anchoring the model learning with trustable examples. This combination of reliable labeled data with a creative use of unlabeled data allows our model to benefit from the full spectrum of available information, leading to more effective and comprehensive learning outcomes.

\subsection*{A. Datasets} 

The MNIST dataset \cite{b11} is a well-known benchmark in the field of machine learning. It consists of 60,000 handwritten digits (0-9), with each digit represented as a $28 \times 28$ pixels grayscale image. This dataset is commonly used for tasks related to digit recognition and image classification, making it a fundamental resource for testing and developing various machine learning algorithms.

The CIFAR-10 dataset \cite{b12} is another widely used dataset in computer vision. It contains 60,000 colored images, divided into 10 classes, with each class representing various everyday objects or animals, such as cars, birds, or cats. These images are relatively small, $32 \times 32$ pixels in size, and serve as a valuable benchmark for testing image classification and deep learning models due to their intrinsic diversity and complexity.

\subsection*{B. Model Architecture}

Our models dedicated to self- and semi-supervised classification on the MNIST and CIFAR-10 datasets use the same general two-part architecture, consisting of an encoder in charge of the features extraction, followed by a Multi-Layer Perceptron (MLP) classifier.

\subsubsection{Encoder}
The encoder architecture is tailored to the target dataset. 

\begin{figure}[h]
\centering
\includegraphics[width=0.5\textwidth]{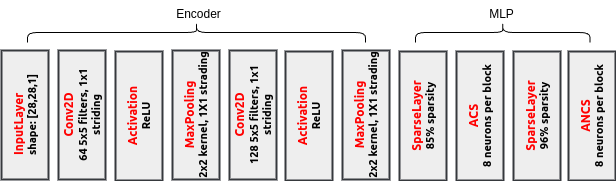}
\caption{Proposed network architecture for MNIST}
\label{fig:MNISTNetwork}
\end{figure}

For MNIST ( see Fig. \ref{fig:MNISTNetwork}): The encoder features two convolutional layers, each followed by a ReLU activation function. At the output of each convolutional layer, max pooling is applied to reduce the feature maps' spatial dimensions, thereby decreasing computational complexity and parameters, and enhancing network efficiency.
% Inserting the image

\begin{figure}[h]
\centering
\includegraphics[scale=0.45]{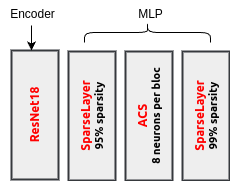}
\caption{Proposed network architecture for CIFAR-10}
\label{fig:CIFAR10Network}
\end{figure}

For CIFAR-10 (Fig. \ref{fig:CIFAR10Network}): To increase learning efficiency for this more challenging dataset, the simple encoder for MNIST is replaced with ResNet-18. ResNet-18 is a deeper Residual Network variant, that incorporates residual connections to facilitate deeper network training, improving image classification performance on complex datasets like CIFAR-10.

\subsubsection{MLP} The architecture of our MLP classifier relies on an innovative structure with two distinct sparse layers and specialized processing blocks. The first sparse layer, inserted after the last max pooling stage, achieves a sparsity level of 85\%, allowing the network to focus on essential features, thus reducing overfitting and improving efficiency. Coming next, the Add-Compare-Select (ACS) function introduces a competition among neurons, activating only the one with the highest value. A second sparse layer follows, with an increased sparsity of 96\%, after which we may find an Add-Normalize-Compare-Select (ANCS) function, but for MNIST only. The ANCS function extends the ACS functionality by incorporating block-by-block normalization. To enable this local normalization, the weights of the final sparse layer are made positive through the utilization of the ReLU activation function.

\section{Experiments}
The performance of the proposed learning architecture has been evaluated on MNIST in a self-supervised learning context, as well as on CIFAR10 in a semi-supervised configuration. 

\subsection{Implementation Details}

To ensure consistent, systematic learning and model refinement for both MNIST and CIFAR-10, we found crucial to organize the training into cycles, epochs, and batches. Cycles ensure complete dataset coverage, epochs allow full iterations over all data batches, and batches enable efficient data processing and incremental model updates, collectively facilitating continuous model improvement. It is important to note that each epoch uses different data augmentation strategies, further enriching the learning and ensuring that the model encounters diverse data representations throughout its training.

For the MNIST dataset, the model architecture is trained through 100 cycles, each cycle consisting of five epochs. At the beginning of each cycle, pseudo-labels estimation is conducted once. These pseudo labels are then used for the subsequent five epochs of training. For training, we use Adam optimizer and a linear learning rate schedule with damping coefficient decaying from 1.0 to 0.001, starting from an initial learning rate of 0.0015. The model's robustness and adaptability are further enhanced by a series of data augmentation techniques: rotation, elastic distortion, random erasing, and center cropping. The Binary Cross-Entropy (BCE) loss is used to measure the difference between predictions and pseudo-labels, paired with a local normalization strategy for block-by-block data processing. This comprehensive approach ensures robust learning and model's effectiveness in discerning complex patterns throughout its extensive training.

Similarly, for CIFAR-10, our model undergoes 300 training cycles, with each cycle comprising five epochs to deeply engage with the complexity of the dataset. At the core of our model is the ResNet18 encoder, selected for its robust feature extraction capabilities. Stochastic Gradient Descent (SGD) with Nesterov momentum is used to optimize the model, starting with an initial learning rate of 0.03. The learning rate is meticulously modulated across the 300 cycles by means of a cosine decay schedule, of the form \(0.03\cos\left(\frac{8\pi s}{16S}\right)\), where \(s\) and \(S\) represents the current and last training steps, respectively. An MSE loss function is used for both labeled and unlabeled batches. For labeled data, the MSE loss quantifies the difference between the model's predictions and the true labels. For unlabeled data, it measures the discrepancy between predictions and generated pseudo-labels. The labeled and unlabeled datasets will be denoted by X and U, respectively. The batch size for X is set to 64. The batch size for U is set to be 8 times the batch size of X, that is $8\times 64 = 512$. In terms of data augmentation, our implementation strictly adheres to the strong and weak augmentation strategies outlined in FixMatch \cite{b13}.

\subsection{Evaluation }

Our comprehensive evaluation strategy for the MNIST dataset relies on two distinct methods to assess the performance of our model using the embeddings as features. Firstly, we use K-means clustering \cite{b14} to categorize the embeddings from the test set into 10 clusters. Each cluster is mapped onto one of the ten digits, based on the majority label of its member embeddings. Secondly, we randomly select one labeled instance from each class in the training set as a representative. We then measure the similarity between the selected instances and the embeddings of the test set to assign class labels. For the CIFAR-10 dataset, the evaluation relies solely on the second method. We choose a single representative labeled example per class, and then assess model performance based on the similarity between the selected examples and the test embeddings. This more focused approach adopted for CIFAR-10 allows us to take advantage of the more complex and varied nature of the dataset, aligning the evaluation strategy with the specific challenges and characteristics inherent to CIFAR-10 images. 

As for the experimental design, we conducted five distinct experiments for both self-supervised and semi-supervised learning methods to confirm the robustness and reliability of our results. Each experiment begins with random initialization of the weights and sparse layers' connections for each network, to avoid any initialization bias and create network diversity. Within each cycle, data augmentation is introduced in a stochastic manner over different sets of training images. Last but not least, to assess the outcomes, we use a dual strategy that combines taking the majority vote from the five models, and computing the average accuracy across different labeled data sets. 

\subsection{Results}

In Table \ref{tab:accuracy_comparison_mnist}, we present the classification accuracy achieved on the MNIST dataset by the proposed self-supervised learning approach. The Table reports the average accuracy across five different labeled data (avg acc), as well as the collective decision-making accuracy obtained by majority vote among the five networks with the same labeled data (maj. vote). Two distinct methodologies were used to assess model performance: one using K-means clustering ('K-means(\%)') and the other leveraging cosine similarity ('cosine sim(\%)') for each network. Together these metrics provide a clear and comprehensive view of the model's performance, demonstrating the effectiveness of both individual networks and a collective ensemble approach in accurately classifying MNIST digits.

\begin{table}[h]
\centering
\caption{Test Accuracy on MNIST with self-supervised learning.}
\label{tab:accuracy_comparison_mnist}
\begin{tabular}{|l|l|l|}
\hline
Method                   & avg acc (\%)\footnotemark[1] & maj. vote (\%)\footnotemark[2] \\ \hline
Fully Supervised \cite{b15}   & 99.83        & 99.87  \\ \hline
self supervised IIC \cite{b16}    & 99.30        &  -    \\ \hline
Ours ( K-means )     & 99.46        & 99.56          \\ \hline
Ours ( Cosine similarity)  & 99.45  & 99.57          \\ \hline
\end{tabular}
\end{table}
\footnotetext[1]{Average accuracy (avg acc): this is the mean of the accuracy values obtained from five networks (five distinct instances of the model).}
\footnotetext[2]{Majority vote (maj. vote): this method aggregates the votes from the five networks to decide the class label.}

Table \ref{tab:accuracy_comparison_cifar10} presents a comparison of the classification accuracy obtained for semi-supervised learning on CIFAR10 with various well-established as well as more recent methods, including the $\Pi$-Model \cite{b9}, Pseudo-Labeling\cite{b18}, Mean Teacher\cite{b19}, UDA\cite{b20}, MixMatch\cite{b21}, FixMatch (RA)\cite{b13}, and Dash\cite{b22}, each referenced accordingly. The Table also showcases the performance of our method, both in terms of average accuracy (avg. acc) and majority vote accuracy (Maj. Vote).
\begin{table}[h]
\centering
\caption{ Test accuracy comparison (mean over 5 runs) for semi-supervised learning on CIFAR-10, using 25 labeled samples per class.}
\label{tab:accuracy_comparison_cifar10}

\begin{tabular}{|l|c|}
\hline
Method & Accuracy (\%) \\ \hline

$\Pi$-Model  & 45.74 \\ 
Pseudo-Labeling  & 50.22 \\ 
Mean Teacher  & 67.68 \\ 
UDA  & 91.18 \\ 
MixMatch   & 88.95 \\ 
FixMatch(RA)  & 94.93 \\ 
Dash  & 95.44 \\  \hline
Our Method  (avg. acc) & 93.51 \\  
Our Method (Maj. Vote) & 94.21 \\ \hline
\end{tabular} 
\end{table}

At this early stage of our research on CIFAR-10 classification, the results obtained with our semi-supervised learning method are quite competitive, achieving an accuracy of 94.21\% with majority vote. What sets our approach apart is the strategic use of sparsity layers, which significantly reduces the number of parameters in the model, thereby increasing its efficiency and speed. Despite the current state-of-the-art for CIFAR-10 being held by Dash with an accuracy of 95.44\%, our method stands out by its promising performance as well as its efficiency. Part of this efficiency gain can be attributed to our use of sparse layers, making our approach an attractive and resource-efficient choice to tackle the CIFAR-10 dataset. The residual accuracy gap of 1.93\% only between our approach and the current leader leaves room for additional improvement, for example by further fine-tuning of hyperparameters and other model enhancements.

\section{Conclusion}
It is possible to design and operate an artificial neural network without having to understand all  its components and behavior in detail. However, there are critical applications, such as autonomous driving or medical diagnostics, which require total control over the explainability of operations performed and decisions made, especially when the network makes mistakes \cite{b23,b24}. One way of ensuring that algorithms behave as desired is to introduce principles and functions that have proved their worth in other technological fields. In this paper, we have compiled a list of concepts from which the telecommunications field has benefited greatly. Among these, it seemed relevant to us to combine redundant coding and spatial diversity in a relatively simple learning architecture integrating competition layers and sparse matrices. The results obtained with self-supervised learning experiments on the MNIST dataset are convincing, with an inference accuracy higher than the previous state-of-the-art. Semi-supervised learning simulations have also been conducted on the more challenging CIFAR-10 dataset. To date, the results are not quite up to the state-of-the-art, yet very close. This is even more promising, as they were obtained without the need for sophisticated mathematical processing. Therefore, our future work will focus on finding the reasons for this performance gap between self-supervised and semi-supervised applications. Throughout our work, we have observed that the classification performance of CIFAR-10 images is very sensitive to the values of hyperparameters, in particular sparsity rates and learning rate, which we have not been able to refine completely. Other avenues could also be investigated: increasing the number of competition-based layers, merging the X and U batches into a single one, replacing binarization (hard decision) with a more progressive function (soft decision), still to be determined.

\vspace{12pt}
\color{red}

\end{document}